\documentclass[review]{elsarticle}
\graphicspath{ {./figures/} }
\usepackage{hyperref}
\usepackage{float}
\usepackage{verbatim} 
\usepackage{apalike}
\usepackage{caption}
\usepackage{subcaption}
\usepackage{tabularx}
\usepackage{adjustbox}
\usepackage{booktabs}
\restylefloat{figure}
\restylefloat{table}
\usepackage{multirow}
\usepackage{graphicx}
\usepackage{xcolor}
\usepackage{amsmath}
\usepackage{datetime2}
\usepackage{enumitem}
\usepackage{todonotes}
\usepackage[noend, ruled,vlined]{algorithm2e}
\usepackage{setspace}

\journal{Expert Systems with Applications}

\biboptions{authoryear}

\begin{document}
\begin{frontmatter}

\begin{titlepage}
\begin{center}
\vspace*{1cm}

\textbf{ \large Federated Learning for Violence Incident Prediction in a Simulated Cross-institutional Psychiatric Setting}

\vspace{1.5cm}

Thomas~Borger$^{a,b}$~(thomasborger2@gmail.com), Pablo~Mosteiro$^a$~(p.mosteiro@uu.nl), Heysem~Kaya$^a$~(h.kaya@uu.nl), Emil~Rijcken$^{c,a}$~(e.f.g.rijcken@tue.nl), Albert~Ali~Salah$^{a,f}$~(a.a.salah@uu.nl), Floortje~Scheepers$^d$~(F.E.Scheepers-2@umcutrecht.nl), Marco~Spruit$^{e,g,a}$~(M.R.Spruit@lumc.nl) \\

\hspace{10pt}

\begin{flushleft}
\small  
$^a$ Department of Information and Computing Sciences, Utrecht University, Utrecht, the Netherlands \\
$^b$ KPMG N.V., Amstelveen, the Netherlands \\
$^c$ Jheronimus Academy of Data Science, Eindhoven University of Technology, ‘s-Hertogenbosch, the Netherlands \\
$^d$ Department of Psychiatry, University Medical Centre Utrecht, Utrecht, the Netherlands \\
$^e$ Department of Public Health \& Primary Care, Leiden University Medical Centre, Leiden, the Netherlands\\
$^f$Department of Computer Engineering, Boğaziçi University, Istanbul, Turkey \\
$^g$Leiden Institute of Advanced Computer Science, Leiden University, Leiden, the Netherlands

\vspace{0.5cm}
\textbf{Corresponding Author:} \\
Pablo Mosteiro \\
Department of Information \& Computing Sciences, Utrecht University, Utrecht, the Netherlands \\
Tel: +1 (609) 759-0645 \\
Email: p.mosteiro@uu.nl

\end{flushleft}        
\end{center}
\end{titlepage}

\title{Federated Learning for Violence Incident Prediction in a Simulated Cross-institutional Psychiatric Setting}

\author[uu,kpmg]{Thomas Borger}
\ead{thomasborger2@gmail.com}

\author[uu]{Pablo Mosteiro~\corref{coraut}}
\ead{p.mosteiro@uu.nl}

\author[uu]{Heysem Kaya}
\ead{h.kaya@uu.nl}

\author[tue,uu]{Emil Rijcken}
\ead{e.f.g.rijcken@tue.nl}

\author[uu,bu]{Albert Ali Salah}
\ead{a.a.salah@uu.nl}

\author[umcu]{Floortje Scheepers}
\ead{F.E.Scheepers-2@umcutrecht.nl}

\author[lumc,liacs,uu]{Marco Spruit}
\ead{M.R.Spruit@lumc.nl}

\cortext[coraut]{Corresponding author.}
\address[uu]{Department of Information and Computing Sciences, Utrecht University, Utrecht, the Netherlands}
\address[kpmg]{KPMG N.V., Amstelveen, the Netherlands}
\address[tue]{Jheronimus Academy of Data Science, Eindhoven University of Technology, ‘s-Hertogenbosch, the Netherlands}
\address[umcu]{Department of Psychiatry, University Medical Center Utrecht, Utrecht, the Netherlands}
\address[lumc]{Department of Public Health \& Primary Care, Leiden University Medical Centre, Leiden, the Netherlands}
\address[bu]{Department of Computer Engineering, Boğaziçi University, Istanbul, Turkey}
\address[liacs]{Leiden Institute of Advanced Computer Science, Leiden University, Leiden, the Netherlands}

\begin{abstract}
\textcolor{black}{Inpatient violence is a common and severe problem within psychiatry. Knowing who might become violent can influence staffing levels and mitigate severity. Predictive machine learning models can assess each patient's likelihood of becoming violent based on clinical notes. Yet, while machine learning models benefit from having more data, data availability is limited as hospitals typically do not share their data for privacy preservation. 
Federated Learning (FL) can overcome the problem of data limitation by training models in a decentralised manner, without disclosing data between collaborators. However, although several FL approaches exist, none of these train Natural Language Processing models on clinical notes.  
In this work, we investigate the application of Federated Learning to clinical Natural Language Processing, applied to the task of Violence Risk Assessment by simulating a cross-institutional psychiatric setting. We train and compare four models: two local models, a federated model and a data-centralised model. Our results indicate that the federated model outperforms the local models and has similar performance as the data-centralised model. 
These findings suggest that Federated Learning can be used successfully in a cross-institutional setting and is a step towards new applications of Federated Learning based on clinical notes. 
}
\end{abstract}


\end{frontmatter}

\section{Introduction}
\label{introduction}


\textcolor{black}{Inpatient violence is a serious problem in clinical psychiatry, causing short- and long-term damage to property as well as people~\citep{VanLeeuwen2017, Inoue2006, Nijman2005, Havaei2019}.}
\textcolor{black}{Violence Risk Assessment (VRA) has been used in mental healthcare to inform medical decisions and mitigation strategies~\citep{Singh2014, Conroy2012}.}
\textcolor{black}{Several }\textcolor{black}{ manual} \textcolor{black}{ VRA methods have been proposed and evaluated~\citep{Almvik2000, Douglas2014, Ogloff2006}, yet these methods are time-consuming and subjective, and some of them require advanced training to use~\citep{Nicholls2006}}.
\textcolor{black}{Machine Learning (ML) methods promise to address these limitations, developing fast and objective predictions based on patient data present in Electronic Health Records (EHR).}

\textcolor{black}{In the psychiatry domain, a particularly promising ML approach is Natural Language Processing (NLP), since EHRs contain large amounts of unstructured clinical notes written by nurses and psychiatrists. The information in these notes could be employed in decision-support systems to aid psychiatrists in predicting aggression, diagnosing patients, predicting side-effects from medication, and predicting suicide attempts, among others. The information is reported in subtle and nuanced ways, and often includes typographical errors, abbreviations and technical terms.}
\textcolor{black}{Not surprisingly, a common problem encountered by ML researchers in the clinical domain are datasets that are small~\citep{Pestian2010} or too specific~\citep{Suchting2018}.}
\textcolor{black}{Thus, increasing dataset size and diversity is desirable for performance of ML models, in particular NLP models used in psychiatry.}

\textcolor{black}{Combining datasets from multiple departments and institutions would be a natural way to enlarge datasets for various tasks. Yet, medical institutions are usually not allowed to combine their data~\citep{umcpht}.}
\textcolor{black}{Thus, instead of sharing data, machine learning models can be shared amongst institutions, using local data for training and/or fine-tuning.}

\textcolor{black}{This is the basis of Federated Learning (FL)~\citep{Mcmahan2016}. Through FL, multiple parties collaborate in solving an ML task under the coordination of a central server,} \textcolor{black}{ where data are never allowed to leave a party's device \citep{Kairouz2021}.} \textcolor{black}{Though some losses are expected with respect to a data-central approach, it has been shown that these could be quite small and acceptable given the gain in privacy~\citep{Sheller2019}. FL has been gaining traction in recent years, and applications within the medical domain are slowly emerging~\citep{Kairouz2021, Deist2020}.} \textcolor{black}{However,} \textcolor{black}{none of the clinical applications of FL so far employ clinical texts.}


\textcolor{black}{In this work, we employ clinical texts for FL, examining violence risk assessment. We seek to find how FL compares to centrally- and locally trained models. For this comparison, } \textcolor{black}{we use free texts in EHRs.}
\textcolor{black}{Since we do not have access to data from multiple institutions, we use ``mock" institutions, created from the data of a single location using nursing-ward-based partitioning.
We train four machine learning models:} \textcolor{black}{  a federated model, a data-centralised model and two local models (A and B). Here, A and B are the names of the mock institutions we created. Then, we compare the performance of these four models on institutions A and B separately and on the combined test dataset.}

\textcolor{black}{Our main contributions are:
\begin{itemize}
    \item We demonstrate that FL applied to NLP models and trained on clinical texts has similar performance as a centralised model, and better than locally trained models.
    \item We highlight the potential of FL for clinical psychiatry.
\end{itemize}
}
The remainder of this paper is structured as follows. Section~\ref{sec:relatedwork} discusses related work regarding FL in the medical domain. Section~\ref{method} describes the dataset, and explains the method for obtaining the empirical results. Sections~\ref{experimental-results} \&~\ref{discussion} state and discuss the empirical results. Finally, Section~\ref{conclusions} provides the conclusions drawn from the results.

\section{Related Work and Background}
\label{sec:relatedwork}

\textcolor{black}{Multiple Machine Learning (ML) methods have been proposed to tackle the problem of Violence Risk Assessment (VRA). \cite{Bader2015} attempted to differentiate between patients perpetrating severe and repeated aggression and non-aggressive patients, using common risk factors as predictor variables. In a retrospective study, \cite{Raja2005} found some factors that seemed to correlate with inpatient violence. \cite{Menger2019}, \cite{Le2018}, and \cite{Cook2016} exploited the abundant free text in EHRs to employ Natural Language Processing (NLP) to this task.}
\textcolor{black}{Beyond VRA, \cite{Pestian2010} used NLP to classify suicide notes as legitimate or elicited.}

\textcolor{black}{Two limiting factors in the development of fair and accurate ML models for the healthcare domain are dataset size and diversity. Of the studies mentioned above, only one had more than a few thousand data points~\citep{Le2018}. Its limitation, however, was that they predicted existing VRA instrument scores, not real violence incidents. \cite{Suchting2018} had nearly 30 thousand data points, yet they report being limited both by dataset diversity (due to the nature of their facility) and by dataset size (due to the imbalanced nature of the dataset, as most patients do not engage in violence).}
\textcolor{black}{Aggregating data from multiple institutions would tackle both problems.} However, as medical data often resides in secure data silos across institutions \citep{Lehne2019}, aggregating these data is not possible.

\textcolor{black}{Federated Learning (}FL\textcolor{black}{)} is a novel technique for training \textcolor{black}{ML} models on decentralised data~\citep{Konecny2016}. It began with the question of how one can train a\textcolor{black}{n ML} model in a setting where data is unevenly distributed across a large number of devices\textcolor{black}{, and the data cannot be shared among devices or with the central server}. 
FL \textcolor{black}{provides a solution to this question} through decentralised training, orchestrated by a central server. The server initialises and sends a model to each participating institution or data silo. Each institution trains the model on their own data, and shares the updated model's parameters with the central server. The server then aggregates all models and creates a new global model. A widely used algorithm for creating a new model is FedAvg \citep{Mcmahan2016}, which performs a weighted average over the parameters of all models to create a new model. Other algorithms have been proposed to allow the use of adaptive optimisers, such as FedAdagrad, FedYogi, and FedAdam \citep{reddi2021adaptive}.

FL has brought promising results in recent literature, where federated models perform nearly on par with data-centralised models for medical classification tasks, such as brain tumour segmentation~\citep{Sheller2019, Li2019} and in-hospital mortality prediction~\citep{Choudhury2019a}. The technique has been applied on private medical data as well by utilising the Personal Health Train (PHT), for classifying post-treatment survival chances in lung cancer patients, by collaborating with eight medical institutions \citep{Deist2020}. PHT is a platform aiming to provide healthcare data from various sources to researchers while ensuring privacy protection. FL has also been used to predict suicidal ideation in online social care texts~\citep{Ji2019}. During the literature search conducted at time of study, no applications of FL on models employing clinical texts were identified. Table~\ref{tab:related_works} shows all the aforementioned methods, together with their goals and limitations.

To bring FL to the psychiatric domain, a \textcolor{black}{N}atural \textcolor{black}{L}anguage \textcolor{black}{P}rocessing~(NLP) task \textcolor{black}{is} chosen for this study. Clinical notes have been written about admitted patients on a daily basis for many years across medical institutions. This means that local datasets are available for research at these institutions. Issues with these clinical texts \textcolor{black}{are} that they are semi-structured, and sometimes contain thousands up till tens of thousands of words for a single admission period, making feature extraction a difficult task. 
\textcolor{black}{\cite{Menger2018} compared various methods to convert texts into vectorial representations, including bag-of-words, TF-IDF, Word2Vec and Doc2Vec. Following previous work~\citep{Mosteiro2021}, in this paper we use Doc2Vec~\citep{le2014distributed}, which generates a fixed-length vector for a piece of text of arbitrary length.} In this study's context, a document is the collection of notes of one admission period of a patient. Through this method, the vector representations aims to keep the semantics within each document intact. The representations can then be fed into an ML model such as a neural network for a classification task.

\begin{table}
    \centering
    \begin{tabular}{l|c|c|c|l}
    \multirow{2}{*}{Paper} & \multirow{2}{*}{FL} & \multirow{2}{*}{NLP} & Clinical & \multirow{2}{*}{Limitations} \\
    &&&texts&\\
    \hline
    \multicolumn{5}{c}{\emph{Violence Risk Assessment}}\\
    \hline
    \cite{Bader2015} &  && & Small cohort \\
    \cite{Raja2005}  & & && Retrospective study \\
    \cite{Menger2019}  & &X& X & Generalisability \\
    \cite{Le2018}  & &X& X & Predicts other instruments \\
    \cite{Suchting2018}  && & & Generalisability \\
    \hline
    \multicolumn{5}{c}{\emph{Suicide Prediction}}\\
    \hline
    \cite{Ji2019} & X &X& & \\
        \cite{Cook2016} & &X& & \\
    \cite{Pestian2010} & &X& X & Small cohort \\
    \hline
    \multicolumn{5}{c}{\emph{Other work}}\\
    \hline
    \cite{Sheller2019} & X & & \\
    \cite{Li2019} & X & & \\
    \cite{Choudhury2019a} & X & & \\
    \cite{Deist2020} & X & & \\
\end{tabular}
    \caption{Previous studies described in Section~\ref{sec:relatedwork}. None of the studies focused on clinical texts employ FL, or vice versa. Limitations are listed, where applicable. \emph{Generalisability} means that the performance reported has been shown or is expected not to generalise to data from new institutions.}
    \label{tab:related_works}
\end{table}

\section{Method}
\label{method}
In this section, we outline the method for conducting the FL experiment for predicting inpatient violence. First the data and the processing steps are described in Section \ref{data} and Section \ref{data_processing}, respectively. Then the setup and training procedure are described in Section \ref{treatment_design}, and the method for validation of the classification models is given in Section \ref{treatment_validation}. Thereafter, more detail is provided regarding the implementation of FL in the experiment in Section \ref{federated_learning_implementation}.

\subsection{Data}
\label{data}

The data made available by the psychiatry ward of UMC Utrecht for this study is the violence incident dataset prepared for violence risk assessment within admitted patients by \cite{Mosteiro2020,Mosteiro2021}. Each data point corresponds to an admission period of a patient, and contains the concatenation of clinical notes of a maximum of 28 days before up until and including the 1st day after admission. 
Based on the next 27 days following the first day of admission, the data points are labelled by whether a violence incident took place or not (positive/negative outcome). 
The clinical notes, which are written in Dutch, have been vectorised using Doc2Vec~\citep{le2014distributed}, with a feature vector dimensionality of 300. 
No structured features such as gender or age were used, as they did not provide significant discriminatory power in previous work~\citep{Mosteiro2020}.
There are four \textcolor{black}{nursing wards in the psychiatry department at the UMC Utrecht}, and each data point belongs to one \textcolor{black}{nursing} ward. The characteristics of the dataset are shown on Table~\ref{table:dataset}.
\begin{table}[H]
 \centering
 \adjustbox{max width=\textwidth}{
    \begin{tabular}{@{}rclccc@{}}\toprule
        \textit{\textcolor{black}{Nursing} ward} & \textit{Age} & \textit{Description} & \textit{Positive} & \textit{Negative} & \textit{Total} \\
        \midrule
        \textit{A1} & $>$40 & \textit{Affective \& psychotic disorders} & 25 & 734 & 759 \\
        \textit{A3} & 15-35 & \textit{Diagnosis \& early psychosis} & 130 & 696 & 826 \\
        \textit{A2V} & $>$18 & \textit{Acute \& intensive care} & 167 & 1710 & 1877 \\
        \textit{A2J} & 12-18 & \textit{Acute \& intensive care} & 103 & 715 & 818 \\
        \midrule
        \textit{Total} & - & -  & 425 & 3855  & 4280 \\
        \bottomrule
    \end{tabular}}
   \caption{Dataset characteristics. Each data point is an \textit{admission period}, i.e., the period that a patient spends while admitted to a given \textcolor{black}{nursing} ward of the psychiatry department. \textit{Age} refers to the age of the patients in the \textcolor{black}{nursing} ward. \textit{Positive} and \textit{Negative} data points are defined by whether the patient is involved in a violence incident during the first 27 days after the first day of the admission period.} \label{table:dataset}
\end{table}

\subsection{Data Processing}
\label{data_processing}

\begin{figure}[ht]
    \centering
    \includegraphics[width=0.70\textwidth]{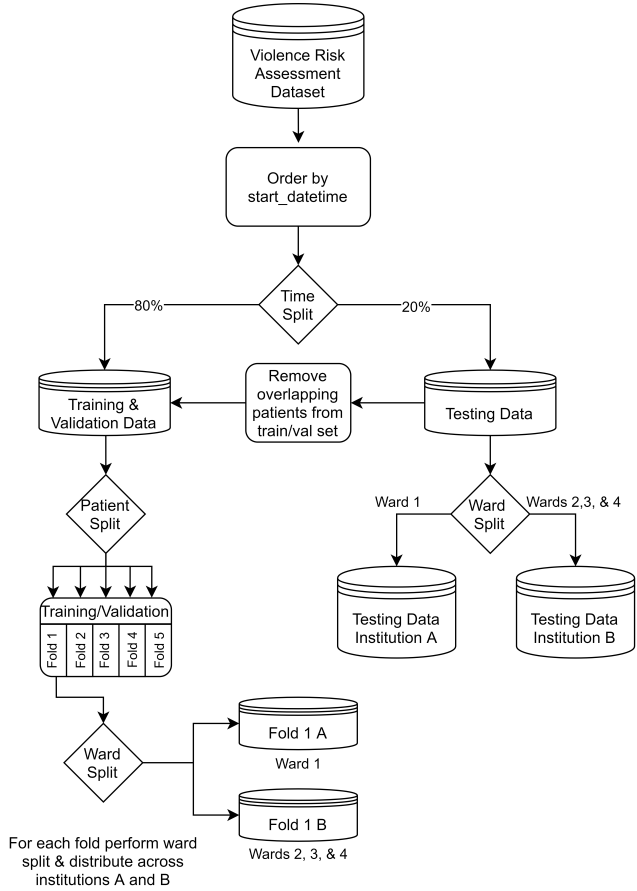}
    \caption[Data Preparation]{The data processing procedure of the violence risk assessment dataset.}
    \label{fig:datapreparation1}
\end{figure}

To simulate two institutions (A \& B) based on one dataset, and to allow for hyper-parameter tuning, a data processing procedure was designed to ensure the split-up datasets meet the following requirements. 
First, each of the four \textcolor{black}{nursing} wards is assigned to either institution A or B, in such a way that makes the numbers of data points in A and B as even as possible. 
Second, both datasets are split up into a train/validation and test set. The train/validation set is split up into 5 folds for cross-validation (CV). Third, between cross-validation folds themselves, and between the train/validation set and the testing set, no patient IDs may overlap; overlapping patient IDs could result in validating/testing on training data. This overlap sometimes occurs when a patient is moved to a different \textcolor{black}{nursing} ward, and the new \textcolor{black}{nursing} ward copies the notes taken from the previous \textcolor{black}{nursing} ward. Fourth, it should be possible to combine the folds between institutions to form patient-independent folds for federated and data-centralised training. Fifth, both testing sets may only include new data based on the admission timestamp, to ensure we test the final models on new data points exclusively. These requirements are visualised as a top-down procedure illustrated in Figure~\ref{fig:datapreparation1}.

\subsection{Treatment Design}
\label{treatment_design}

In this study, four treatments are designed and compared to test all scenarios derived from the research goals mentioned in Section~\ref{introduction}, based on Wieringa's design cycle~\citep{Wieringa2014}. Each treatment performs a grid search with 5-fold cross-validation (CV) to search for the best hyper-parameters for training a neural network on their respective dataset. Based on this outcome, each treatment delivers a final model by training on the data from all five folds, and is tested against a held-out testing set. These four final models are compared as part of the statistical difference-making experiment.

The difference between each treatment lies within the data it's applied on, and the training method. Two treatments are trained on data from the two simulated institutions A and B. The other two treatments, data-centralised and federated, train on data from both institutions. The data-centralised treatment trains on all data without restrictions, to show how performance would be if privacy regulations could be ignored. Therefore, it acts as a gold standard in terms of performance, as we expect the nonrestrictive training environment to deliver the best performance. The federated treatment trains a neural network on both institutional data sets through FL.

\subsubsection{Classification Model}
\label{classification_model}
The classification model used across treatments is a feed-forward neural network, consisting of an input layer, one hidden layer, and output layer. The size of the input layer corresponds to the number of elements in the Doc2Vec vectors in the dataset (300). The hidden layer size is given by the variable $h$, whose values are optimised through hyper-parameter tuning. Furthermore, the hidden layer uses the Rectified Linear Unit (ReLU) activation function, chosen for its fast computation time. The output layer has a single neuron with a sigmoid activation function for providing the classification.

The model uses the Binary Cross Entropy (BCE) with logit loss function to compute its gradients. We use mini batch gradient descent. Equation~\ref{eq:lossfn} is the average loss per data point for a single batch $n$, given input $x$ and outcomes $y$.
\begin{align}
\label{eq:lossfn}   l(x,y)_n = \frac{1}{T_n} \sum_{i=1}^{T_n}  -[p\,y_i \cdot \log(\sigma(x_i)) &+ (1 - y_i) \cdot \log(1-\sigma(x_i))] 
\end{align}
Batch $n$ contains $T_n$ data points. For each data point $i$, we apply a sigmoid activation function $\sigma$ to the input $x_i$. 
To mitigate the issue of class imbalance, when the binary outcome $y_i$ is positive, we multiply it by a weight $p$ equal to the ratio of negative to positive samples in the dataset.

An exponential learning rate scheduler is used for training, which updates the learning rate through $lr = lr_0 \cdot \gamma ^{n_e}$, where $lr_0$ is the starting learning rate, $\gamma$ is the amount of decay, and $n_e$ is the number of the current epoch. A $\gamma$ of $0.975$ is used for the experiment. This value causes the learning rate to approximately be divided by 10 at epoch 100. It is a relatively quick drop, but as there is a computational constraint in the amount of epochs we can use, we aim for models with an initial quick convergence, and use the remaining epochs for more fine-grained model updates.
The maximum number of epochs is 120.

An early stopping mechanism keeps track of the validation loss of the model at each epoch. The mechanism saves a checkpoint of the model whenever the validation loss decreases since the last overall decrease in validation loss. It means that if the validation loss has not decreased in the last seven epochs, the early stopping mechanism kicks in and stops the training. It will then load the model checkpoint which has the lowest validation loss. This checkpoint model is then used for model evaluation.

\subsubsection{Hyper-parameter Tuning}

\begin{figure}[!htb]
    \centering
    \includegraphics[width=\textwidth]{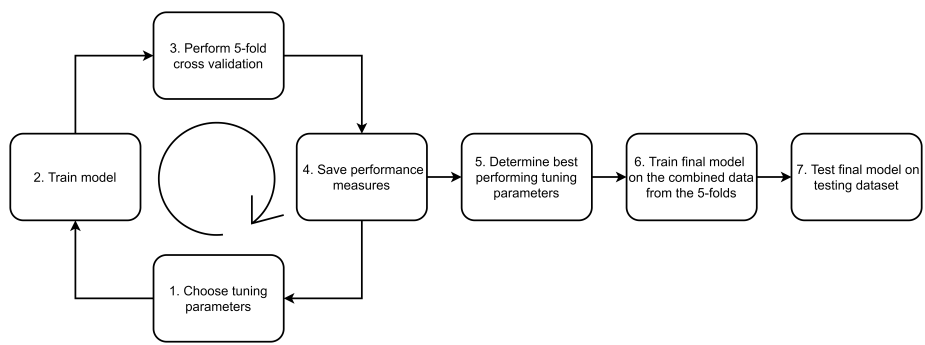}
    \caption[Hyper-parameter tuning cycle]{The hyper-parameter tuning cycle of each treatment.}
    \label{fig:parametertuningcycle}
\end{figure}

Each treatment follows the hyper-parameter tuning cycle and testing procedure as shown in Figure \ref{fig:parametertuningcycle}. This aims to result in hyper-parameter values optimal for training a treatment's final model. The tuning happens through a process known as grid search in steps 1 through 4 in the Figure, where for each possible combination from a fixed set of hyper-parameters, a model is trained to reveal its respective performance. To ensure a good error estimate, the grid search is performed through 5-fold CV. Thus, for each hyper-parameter combination, five models are trained. To compute a performance measure of a combination, the ground truth labels and the predicted labels from the five models are concatenated and used as input for performance measure calculations.

The following set of hyper-parameters are used during the grid search, resulting in 36 unique combinations. These values were chosen during data exploration.
\begin{itemize}
    \item{Hidden Layer Size: [64, 128, 256, 512]}
    \item{Learning rate: [0.005, 0.001, 0.0005]}
    \item{Weight decay: [1e-3, 1e-4, 1e-5]}
\end{itemize}

The performance measure used for fine-tuning is the F1-score, which assigns importance to correctly classifying the positive class. As the violence risk assessment dataset is strongly imbalanced and as correctly classifying patients exhibiting violence is deemed to be more important than correctly classifying patients who do not exhibit such behaviour, being able to accurately evaluate true positives among the positive predictions is key. When the combination with the best F1-score is determined in step 5, the final model can be trained. It uses the best hyper-parameters to train on the data from all 5-folds in step 6. After training, it is ready to be evaluated on the held out testing data at step~7.

\subsection{Treatment Validation}
\label{treatment_validation}

Each treatment is validated by testing its final model on the held out testing data containing only new data points based on the admission timestamp. It corresponds to the final step in the the hyper-parameter tuning cycle (Figure \ref{fig:parametertuningcycle}). This is done by feeding the testing data from institutions A, B, and the combination of the two sets into each treatment's final model. From here, performance measures per treatment are computed. As each treatment uses the same set of testing datasets, and each data point is fed in an identical order to each treatment, the performance measures as well as the raw predictions can directly be compared.

For each performance measure, confidence intervals are calculated by bootstrapping the test set for 10\,000 bootstraps. Bootstrapping is a method for estimating the sample distribution for a certain statistic. It is implemented by sampling the test set with replacement to produce a new test set with a different distribution of the same size, this is done 10\,000 times. From here a bootstrapped mean and confidence intervals can be calculated. The 95\% confidence intervals are computed for each performance measure, treatment, and test set combination. These intervals are computed through percentiles, which is known as the percentile bootstrap method, and are compared between treatments and test sets, to provide insights into relative performance.

Each treatment's bootstrapped performance measures are compared against the federated measures by calculating the difference between the measure scores for each bootstrapped sample. Computing this difference is a method adapted from \cite{cumming2005}. This too will yield a distribution with 95\% confidence intervals. 
If the confidence interval whiskers exclude zero, then the difference is statistically significant. An important side-note for this method is that bootstrapping is ideally performed on the training set. Due to computational limitations, we performed it on the test set to see how much our specific trained models vary in their performance, when the test set is modified slightly through bootstrapping.

\subsection{Federated Learning Implementation}
\label{federated_learning_implementation}

The Python library PySyft is used for simulating a federated setting on a single device. We simulate two institutional devices and a central server. The training/validation algorithm implemented using PySyft is shown below in Algorithm~\ref{alg:fl-implementation}. First, a central model is initialised on the server and a copy is sent to both institutions. Each institution trains the model in batches of their full local dataset. For each batch passing through the institutional models, the models are updated accordingly. After a single pass over the full dataset, the resulting institutional models are sent to the central server. Here they are aggregated by averaging the weights and biases of each layer in the neural network. The resulting averaged model is validated at each epoch to track the validation loss and performance measures. This happens by sending a copy of the averaged model back to the institutions and by feeding it the local validation sets. This results in a set of predictions and ground truth values from both institutions, which are in turn shared with the central server. The central server then computes the performance measures over the concatenation of the predicted and ground truth values from both institutions. 
\begin{algorithm}[H]
\DontPrintSemicolon
\SetAlgoLined
\setstretch{1}
\SetKwFunction{Fvalidate}{validate}
\SetKwFunction{Flocaltrainmodel}{local\_train\_model}
\SetKwFunction{Flocalvalidatemodel}{local\_validate\_model}
\SetKwProg{Fn}{Function}{:}{}
\KwResult{Model, performance}\;
 model = initialise\_model()\;
 \For{$n_{e}$ {\normalfont \textbf{in}} n\_epochs}{
    \For{inst {\normalfont \textbf{in}} institutions}{
        updated\_model\_inst = local\_train\_model(model, $inst$)\;
        }
    model = average(\{updated\_model\_inst\})\;
    (performance, loss) = validate(model)\;
    should\_stop = early\_stopping(loss) \tcp*{Section \ref{classification_model}}\;
    \uIf{\normalfont should\_stop}{
        break\;
    }
}
\Return model, performance\;
\

\Fn{\Fvalidate{{\normalfont model}}}{
        \For{inst {\normalfont \textbf{in}} institutions}{
        predictions\_inst, labels\_inst = local\_validate\_model(model, $inst$)\;
        }
    predictions = concatenate(\{predictions\_inst\})\;
    labels = concatenate(\{labels\_inst\})\;
    performance, loss = get\_performance\_and\_loss(predictions, labels)\;
    \KwRet (performance, loss)\;
  }\;

\Fn{\Flocaltrainmodel{{\normalfont model}, $inst$}}{
    updated\_model = model\;
    \For{$n_b$ {\normalfont \textbf{in}} n\_batches}{
        updated\_model = train(updated\_model, dataset($inst$))\;
        }
    \Return updated\_model\;
}\;

\Fn{\Flocalvalidatemodel{{\normalfont model}, $inst$}}{
    predictions, labels = model(dataset($inst$))\;
    \Return (predictions, labels)\;
}\;

\caption{Federated Learning implementation}
\label{alg:fl-implementation}
\end{algorithm}

\section{Experimental Results}
\label{experimental-results}

\subsection{Data Splitting}
\label{data-splitting}

The violence risk assessment dataset contains a total of 4005 data points after removing overlapping patients from the training/validation set. The first split assigns 856 points to the testing set, and 3149 to the training/validation set. The test set is distributed to institutions A \& B based on \textcolor{black}{nursing} wards, giving institution A 347 and institution B 509 testing data points. Distributing the training/validation set based on the same \textcolor{black}{nursing} ward split assigns 1410 (of which 114 positive) and 1739 (of which 184 positive) data points to institutions A and B, respectively. Table \ref{table: griddata} displays the distribution across the cross-validation folds illustrating the class imbalance based on the training/validation set. \textcolor{black}{Nursing} ward A2V is appointed to institution A, and \textcolor{black}{nursing} wards A1, A2J, and A3 to institution B.

\begin{table}[H]
 \centering
 \adjustbox{max width=\textwidth}{
    \begin{tabular}{@{}rccccccccc@{}}\toprule
        \textit{Treatment} & \phantom{abc} &  \multicolumn{2}{c}{\textit{Institution A}} & \phantom{abc} & \multicolumn{2}{c}{\textit{Institution B}} & \phantom{abc} & \multicolumn{2}{c}{\textit{Combined}}\\
        \cmidrule{3-4} \cmidrule{6-7} \cmidrule{9-10}
        \textit{Label} && \textit{Negative} & \textit{Positive} && \textit{Negative} & \textit{Positive} && \textit{Negative} & \textit{Positive}\\ 
        \midrule
        \textit{Fold 1} && 257 & 26 && 313 & 34 && 570 & 60\\
        \textit{Fold 2} && 277 & 22 && 295 & 36 && 572 & 58\\
        \textit{Fold 3} && 265 & 18 && 311 & 36 && 576 & 54\\ 
        \textit{Fold 4} && 263 & 17 && 307 & 43 && 570 & 60\\
        \textit{Fold 5} && 234 & 31 && 329 & 35 && 563 & 66\\
        \midrule
        \textit{Total} && 1296 & 114  && 1555  & 184 && 2851 & 298 \\
        \bottomrule
    \end{tabular}}
   \caption[Cross-validation data split results]{The distribution of positive and negative data points across institutional cross-validation folds.} \label{table: griddata}
\end{table}

\subsection{Grid Search Cross Validation}
\label{grid_search_cross_validation}

Table \ref{table: gridcv} displays the relationship between cross-validation F1-scores and the F1-scores of applying the treatments to the held-out test set. The goal of the grid search is to find the combination giving the highest F1-score (CV Max F1), and in this way it aims to find a combination which has a comparably high F1-score on the held out testing set. For the final models of the data-centralised, federated, and institution A treatments, the F1-score on their own test set is higher than the CV Max F1-score. This indicates that the hyper-parameters picked during cross validation provides a decent F1-score on the held out test set. Only for institution B the opposite was true as the F1-score on its own test set is lower. In an ideal situation, a similar F1-score is preferred as the cross-validation would then provide the most realistic carry-over value. 

\begin{table}[H]
  \centering
    \footnotesize
    \begin{tabular}{@{}rrrrr@{}}
    \toprule
    \textit{Treatment} & \textit{CV Min F1} & \textit{CV Mean F1} & \textit{CV Max F1} & \textit{F1 own Test Set} \\ \midrule
      \textit{Institution A}     & 0.220 & 0.288 & 0.320 & 0.351  \\
      \textit{Institution B}      & 0.351 & 0.374 & 0.388 & 0.335  \\
      \textit{Federated} & 0.334 & 0.346 & 0.362 & 0.388  \\
      \textit{Data-centralised}          & 0.324 & 0.343 & 0.359 & 0.396  \\ 
      
    \bottomrule
    \end{tabular}
   \caption[Grid CV F1 results]{Grid search cross-validation F1-scores compared to the F1-score on a treatment's own testing set.} \label{table: gridcv}
\end{table}

\subsection{Performance measures}
\label{performance-measures}

We observe in Table \ref{table: performance measures} that the federated and data-centralised models perform much alike regardless of the performance measure on our testing set. This indicates that the FL process has had no large impact on the performance of the model. When comparing the federated model to the local institutional models, we see a large gap in terms of the F1-score for all testing sets; the federated model's F1-score was remarkably higher for each set. In addition, the federated model achieves higher scores for most performance measures regardless of the testing set compared to the local models, while achieving similar scores compared to the data-centralised model.
\textcolor{black}{The data-centralised ROC-AUC score on the combined test set is consistent with that found in previous work~\citep{Mosteiro2021}.}

\begin{table}[H]
  \centering
  \begin{adjustbox}{width=\textwidth,center}
    \begin{tabular}{@{}rrrrrcrrrrcrrrr@{}}\toprule
    \textit{Test set} & \multicolumn{4}{c}{\textit{Test set Institution A}} & \phantom{abc}& \multicolumn{4}{c}{\textit{Test set Institution B}} &
    \phantom{abc} & \multicolumn{4}{c}{\textit{Test set Combined}}\\
    \cmidrule{2-5} \cmidrule{7-10} \cmidrule{12-15}
    \textit{Treatment} & \textit{Inst A} & \textit{Inst B} & \textit{Fed} & \textit{DC} && \textit{Inst A} & \textit{Inst B} & \textit{Fed} & \textit{DC} && \textit{Inst A} & \textit{Inst B} & \textit{Fed} & \textit{DC}\\ \midrule
      \textit{ROC-AUC}     & \textbf{0.803} & 0.744 & 0.777 & 0.774  &&  0.740 & 0.755 & 0.759 &\textbf{0.762} &&  0.764 & 0.742 & \textbf{0.765} & \textbf{0.765}\\
      \textit{PR-AUC}      & 0.278 & 0.274 & \textbf{0.293} & 0.281  &&  0.276 & 0.293 & 0.292 & \textbf{0.296}  &&  0.270 & 0.281 & \textbf{0.288} & \textbf{0.288}\\
      \textit{F1}          & 0.351 & 0.339 & \textbf{0.419} & 0.417  &&  0.325 & 0.335 & 0.366 & \textbf{0.382}  &&  0.336 & 0.337 & 0.388 & \textbf{0.396}\\
      \textit{Recall}      & 0.459 & \textbf{0.757} & 0.703 & 0.676  &&  0.306 & 0.516 & 0.516 & \textbf{0.532}  &&  0.364 & \textbf{0.606} & 0.586 & 0.586\\
      \textit{Precision}   & 0.283 & 0.219 & 0.299 & \textbf{0.301}  &&  \textbf{0.345} & 0.248 & 0.283 & 0.297  &&  \textbf{0.313} & 0.233 & 0.290 & 0.299\\
    \bottomrule
    \end{tabular}
   \end{adjustbox}
   \caption[Experiment performance measures]{Performance measures for each treatment tested on each testing set. Values in bold are the highest among a measure given a specific test set across the four treatments. F1-score, recall, and precision use a classification threshold of 0.5.} \label{table: performance measures}
\end{table}

\subsection{Bootstrapped F1-scores comparison}

\begin{figure}[htpb]
     \centering
     \begin{subfigure}[b]{0.9\textwidth}
         \centering
         \includegraphics[width=0.9\textwidth]{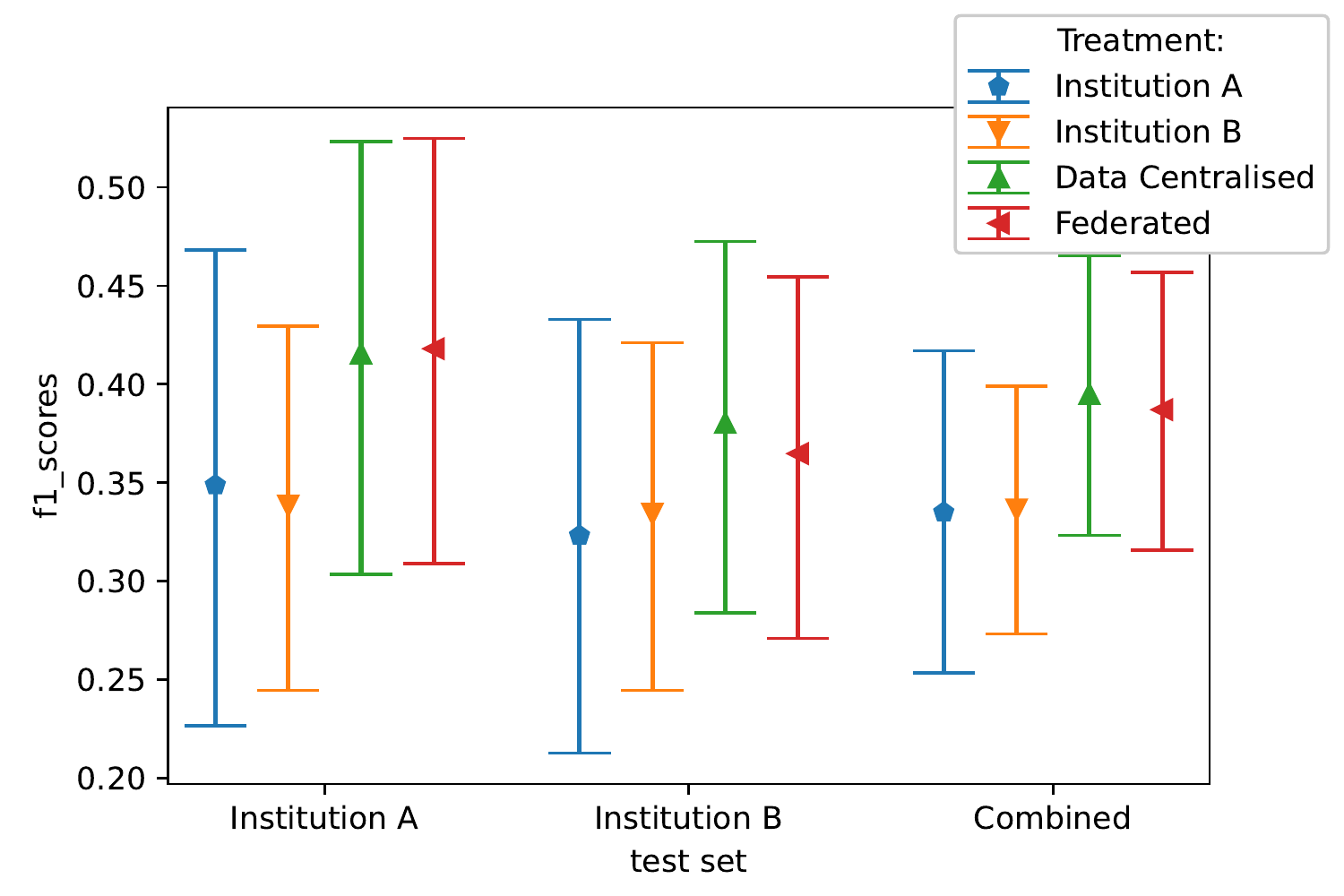}
         \caption{F1-scores CI}
         \label{fig:f1-ci-a}
     \end{subfigure}
     \hfill
     \begin{subfigure}[b]{0.9\textwidth}
         \centering
         \includegraphics[width=0.9\textwidth]{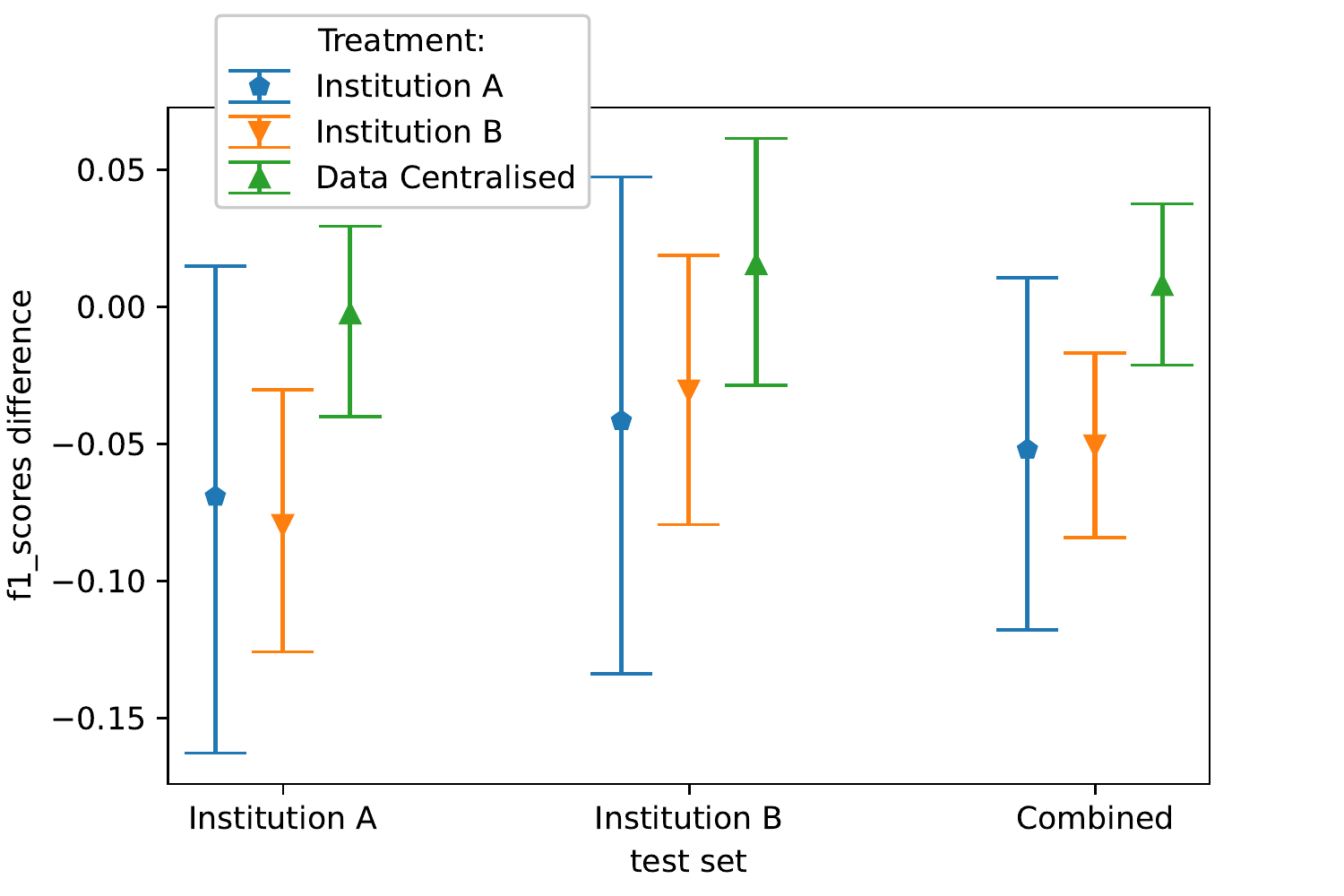}
         \caption{F1-scores differences compared to Federated}
         \label{fig:f1-ci-b}
     \end{subfigure}
    \caption{Comparison of Confidence Intervals (95\%) based on bootstrapped F1-scores. The x-axis refers to the test set. The center points represent the mean of the bootstrapped F1-scores.}
    \label{fig:f1-ci}
\end{figure}

For a given performance measure, the confidence intervals are calculated in two ways. Both methods rely on bootstrapping of the ground truth labels and the predictions based on 10\,000 resamplings of the testing sets. The first method computes a performance measure for each bootstrapped sample. This results in a distribution of a given measure's scores with 10\,000 data points. Then the two-tailed confidence intervals are calculated using percentiles (CI: 95\%). The confidence intervals of this method using the F1-score as a performance measure, are illustrated in Figure \ref{fig:f1-ci-a}. The second method compares the bootstrapped distributions of the F1-scores between all non-federated treatments compared to the federated treatment. Given two treatments, the difference in a performance measure for each bootstrapped sample is calculated. These differences for a given measure provide a new distribution for which the confidence intervals are calculated (CI: 95\%). This method is illustrated in Figure \ref{fig:f1-ci-b}.

\subsection{Prediction comparisons}
\label{prediction_comparisons}
To provide a more in-depth comparison between each treatment's predictions, the confusion matrices and contingency tables are displayed in Tables \ref{table:confusion} and \ref{table:contingency}, respectively. We observe significant differences between the predictions of the two local models and the data-centralised and federated model. Comparing the predictions of the data-centralised and federated models alone, reveals highly similar predictions. Table \ref{table:common} provides the extent to which all models agree on each test sample. We observe that all models often agree with one another on the same test samples. An interesting observation is that the models also commonly misclassify a significant number of similar test samples.

\begin{table}[H]
    \centering
    \subfloat[\centering Institution A ]{ 
        \begin{tabular}{cc|cc}
        \multicolumn{2}{c}{}
                    &   \multicolumn{2}{c}{Predicted} \\
            &       &   Neg &   Pos              \\ 
            \cline{2-4}
        \multirow{2}{*}{\rotatebox[origin=c]{90}{Actual}}
            & Neg    & 678   & 79                 \\
            & Pos    & 63    & 36                \\ 
            \cline{2-4}
            \end{tabular}
         }\quad
    \subfloat[\centering Institution B ]{
        \begin{tabular}{cc|cc}
        \multicolumn{2}{c}{}
                    &   \multicolumn{2}{c}{Predicted} \\
            &       &   Neg &   Pos              \\ 
            \cline{2-4}
        \multirow{2}{*}{\rotatebox[origin=c]{90}{Actual}}
            & Neg    & 560   & 197                 \\
            & Pos    & 39    & 60                \\ 
            \cline{2-4}
            \end{tabular}
         }\\

    \subfloat[\centering Federated ]{ 
        \begin{tabular}{cc|cc}
        \multicolumn{2}{c}{}
                    &   \multicolumn{2}{c}{Predicted} \\
            &       &   Neg &   Pos              \\ 
            \cline{2-4}
        \multirow{2}{*}{\rotatebox[origin=c]{90}{Actual}}
            & Neg    & 615   & 142                 \\
            & Pos    & 41    & 58                  \\ 
            \cline{2-4}
            \end{tabular}
         }\quad
    \subfloat[\centering Data-centralised ]{ 
    \begin{tabular}{cc|cc}
    \multicolumn{2}{c}{}
                &   \multicolumn{2}{c}{Predicted} \\
        &       &   Neg &   Pos              \\ 
        \cline{2-4}
    \multirow{2}{*}{\rotatebox[origin=c]{90}{Actual}}
        & Neg    & 621   & 136                 \\
        & Pos    & 41    & 58                \\ 
        \cline{2-4}
        \end{tabular}
     }
    \caption[Confusion Matrices]{Confusion matrices of each treatment applied to the combined test set.} 
    \label{table:confusion}
\end{table}

\begin{table}[H]
    \centering
    \subfloat[\centering Institution A compared to federated]{ 
        \begin{tabular}{cc|cc}
        \multicolumn{2}{c}{}
                    &   \multicolumn{2}{c}{Institution A} \\
            &       &   Correct &   Incorrect              \\ 
            \cline{2-4}
        \multirow{2}{*}{\rotatebox[origin=c]{90}{Fed}}
            & Correct      & 647   & 26                 \\
            & Incorrect    & 67    & 116                \\ 
            \cline{2-4}
            \end{tabular}
         }
    \subfloat[\centering Institution B compared to federated]{ 
        \begin{tabular}{cc|cc}
        \multicolumn{2}{c}{}
                    &   \multicolumn{2}{c}{Institution B} \\
            &       &   Correct &   Incorrect              \\ 
            \cline{2-4}
        \multirow{2}{*}{\rotatebox[origin=c]{90}{Fed}}
            & Correct      & 608   & 65                 \\
            & Incorrect    & 12    & 171                \\ 
            \cline{2-4}
            \end{tabular}
         }
         \\
    \subfloat[\centering Data-centralised compared to federated]{ 
        \begin{tabular}{cc|cc}
        \multicolumn{2}{c}{}
                    &   \multicolumn{2}{c}{Data-centralised} \\
            &       &   Correct &   Incorrect              \\ 
            \cline{2-4}
        \multirow{2}{*}{\rotatebox[origin=c]{90}{Fed}}
            & Correct      & 658   & 15                 \\
            & Incorrect    & 21    & 162                \\ 
            \cline{2-4}
            \end{tabular}
         }
    \caption[Contingency Tables]{Contingency tables of each treatment applied to the combined test set compared to the federated model.} 
    \label{table:contingency}
\end{table}

\begin{table}[H]
    \footnotesize
    \centering
    \begin{tabular}{cc|cc}
    \multicolumn{2}{c}{}
                &   \multicolumn{2}{c}{} \\
        & Ground Truth      &   Misclassified &   Correct              \\ 
        \cline{2-4}
    \multirow{2}{*}{\rotatebox[origin=c]{90}{}}
        & Neg (total nr: 757)    & 72   & 545                 \\
        & Pos (total nr: 99)  & 34    & 33                \\ 
        \cline{2-4}
        \end{tabular}
    \caption[Common (mis-)classifications]{Commonly correct and commonly misclassified predictions. Common is defined as having all models agreeing upon a single outcome.} 
    \label{table:common}
\end{table}

\begin{figure}[H]
  \makebox[\linewidth]{
    \subfloat[t-SNE]{\label{fig12:a}\includegraphics[width=0.5\linewidth]{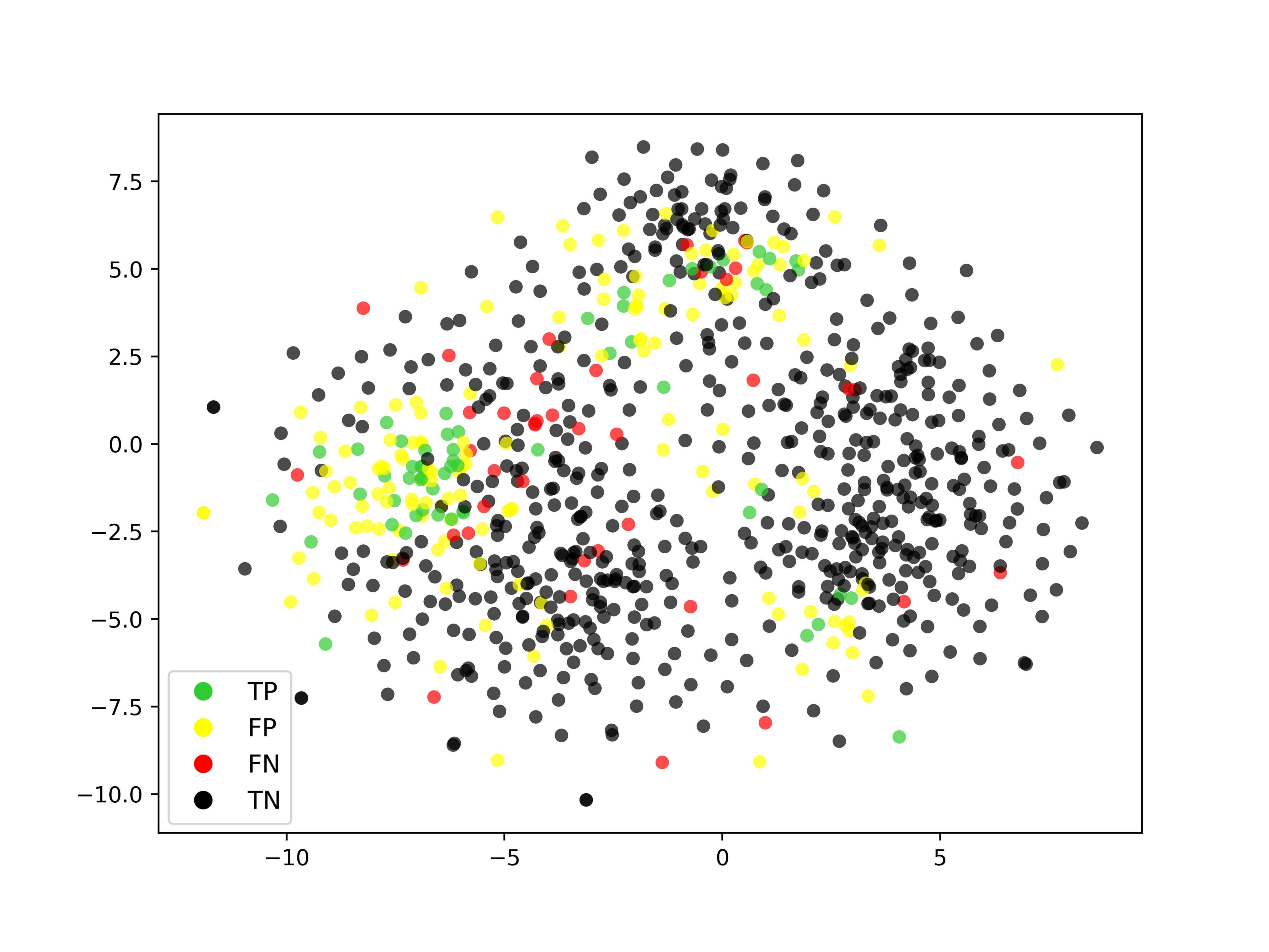}}
    \subfloat[First two principal components]{\label{fig12:b}\includegraphics[width=0.49\linewidth]{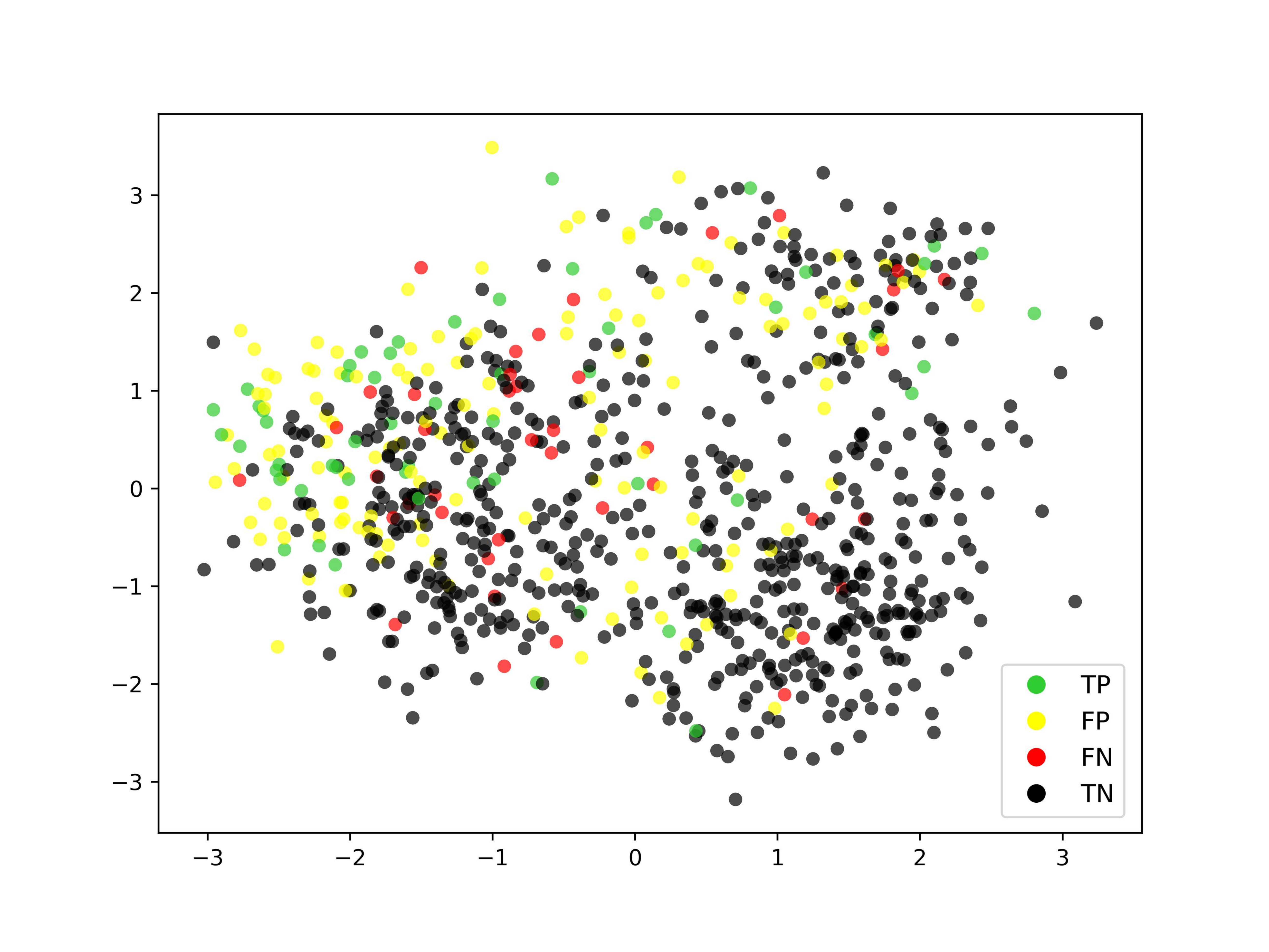}}%
    }
    \caption[PCA and t-SNE on combined test set]{PCA and t-SNE visualisations on the combined test set. The data points are labelled by True Positives (TP), False Positives (FP), False Negatives (FN), and True Negatives (TN), as predicted by the federated model.}
    \label{fig:pcat-snetest}
    
\end{figure}

To provide more insights into the data structure and the classifications, t-SNE and PCA analyses were performed on the testing dataset of the federated model. Figure \ref{fig:pcat-snetest} shows the result of these analyses coloured by the classification of the federated model. It reveals the difficulty of the classification task as both the t-SNE and PCA show that the true positive samples are scattered across the figures. 

\section{Discussion}
\label{discussion}

\subsection{Statistical significance}
\label{Statistical significance}
When tested on the combined testing data, the federated model achieved an F1-score of 0.388 and the data-centralised achieved 0.397. This is in line with our expectations that both models would perform on par with one another. 
We also report on Table~\ref{table: performance measures} the areas under the Receiver Operating Characteristic (ROC-AUC) and precision-recall curve (PR-AUC). These measures indicate performance without choosing a classification threshold.
On the combined test set, both models had the same ROC-AUC (0.765) and PR-AUC (0.288). Furthermore, there was not enough evidence to reject the null hypothesis that the data-centralised and federated model were similar, with a significance threshold of 0.05. While we cannot conclusively state that the federated and data-centralised model perform on par, they did so on our test dataset.

The F1-score of institution A, as observed on its own test set (0.351), was much lower than the federated model (0.419). Even though the observed differences from the experiment were large, the bootstrapped confidence intervals could not prove a statistically significant difference, given a significance threshold of 0.05. However, we can state that the federated model was better on our testing data. Analysing the F1-scores of institution B on its own test set (0.335) compared to the federated model (0.366), we see a smaller difference. For institution B, we observed a statistically significant difference when tested on the testing set from institution A and the combined set, but not when tested on the set from institution B. Again, based on the bootstrapped mean F1-scores we observe that the federated model outperformed the local model, but we cannot state with statistical significance that federated model is better.

\subsection{Model Differences}
\label{model_differences}
The confusion matrices of all models as displayed in Table~\ref{table:confusion} reveal the differences in correct and incorrect predictions between each model. Institution A delivered the most accurate model giving the lowest number of incorrect predictions ($79 + 63 = 142$). However, the model also yielded the highest number of false negatives ($63$) and lowest number of true positives ($36$), which is why it had the lowest F1-score out of all models. The data-centralised model and federated model give an identical number of false negatives ($41$) and true positives~($58$), while differing only by six samples for true negatives and false positives. The model of institution B yielded the highest number of true positives ($60$) at the expense of the highest number of false positives ($197$). The differences between institutions A and B remind us that the F1-score is not the end of the story. It will also depend on whether a practitioner favours a high number of false negatives over a high number of false positives, or vice versa.

Contingency tables reveal the relative performance and agreement of each model compared to the federated model on the combined data. Tables \ref{table:contingency}a \& \ref{table:contingency}b show that there is a significant disagreement between the two local models compared to the federated model. Table \ref{table:contingency}c shows that the highest level of agreement is between the data-centralised and federated model, disagreeing only on 36 ($21 + 15$) data points. 

To see the extent to which all models agree with one another, Table~\ref{table:common} displays the number of commonly correctly classified and the commonly misclassified test samples. Common, within this context, is defined as a prediction for which all models agree with one another on a given test sample. Out of all test samples ($757 + 99 = 856$), overall common agreement ($72 + 33 + 545 + 34 = 684$) between models was as high as 79.9\%. A worrying detail is that all models misclassified 34 real positive samples. This high number of common false negatives begs the question of whether these data points have anything in common as to be classified incorrectly. 

\subsection{Limitations}
\label{limitations}
The first limitation of this study is concerning the Doc2Vec model. The model was trained on all clinical notes present in the violence risk assessment dataset. There are two issues with this approach. First, it would have been more realistic to train this Doc2Vec model using FL; currently it was trained with a data-centralised approach to create the best Doc2Vec model we could with the limited size of the data set. For a real life scenario, adding federated Doc2Vec training to the pipeline is a prerequisite, for the same reasons data-centralised training is not allowed cross-institutionally. However, Doc2Vec is not compatible with PySyft at the moment. The second issue is that the Doc2Vec model has also been trained on clinical notes occurring in the testing set. When making real life predictions, it is unusual to retrain the Doc2Vec model to include the testing clinical notes and retrain the classification model. Rather, one would use the existing Doc2Vec model to immediately vectorise the newly acquired clinical notes to predict for violence. This limitation will be addressed in future work.

Another limitation is that other privacy preserving technologies were not investigated in this study. Models trained through FL can be attacked like other machine learning models, and an attacker might be able to infer details about the training data of the model. FL can be combined with other privacy preserving techniques such as differential privacy \citep{dwork2006}, homomorphic encryption \citep{Gentry2009}, and secure multiparty computation \citep{Yao1982}. These techniques might alleviate additional privacy concerns, but could also negatively impact model performance. To guarantee a high level of privacy to admitted patients, combining FL with these techniques might be required.

\textcolor{black}{Lastly, we observed a large variance in the performance measures and believe this can be attributed to the small test set with a low number of positive samples. Because of this, there is a high probability that a bootstrapped sample contains a skewed class distribution, which has a high impact on the variance of F1-scores.
}

\section{Conclusions}
\label{conclusions}

\textcolor{black}{Violence Risk Assessment (VRA), like many other clinical tasks, can be tackled with Machine Learning methods. In the psychiatry domain, NLP methods are particularly interesting thanks to the abundance of clinical notes containing valuable information. NLP models benefit enormously from bigger and more diverse datasets, as can be acquired by working across multiple institutions. Since data sharing among institutions is not possible, we have developed a Federated Learning (}FL\textcolor{black}{) pipeline} for training a\textcolor{black}{n algorithm} for VRA. \textcolor{black}{We found no performance loss from using FL, as opposed to a data-centralised approach. Also, FL seems to improve the performance of locally trained models tested on a different dataset. To the best of our knowledge, this is the first application of FL and NLP on clinical texts.}

 The results suggest that there are benefits to using federated models and this should be investigated further with cross-institutional datasets. Not only would this provide insights into real-life deployment, it would also lead to more data points for training and testing and could help to decrease performance measure variance.

\textcolor{black}{In future work, we plan to train document embeddings in a federated environment. Furthermore, we will investigate how FL can help solve other clinical tasks, such as text de-identification. Finally, we plan to investigate the possibility of adding other additional privacy-preserving technologies, such as differential privacy.}
\section*{CRediT authorship contribution statement}

\textbf{Thomas Borger:} Methodology, Software, Validation, Formal analysis, Investigation, Data Curation, Writing - Original Draft, Visualization. \textbf{Pablo Mosteiro:} Methodology, Software, Investigation, Data Curation, Supervision, Project administration, Writing - Review \& Editing. \textbf{Heysem Kaya:} Methodology, Supervision, Writing - Review \& Editing. \textbf{Emil Rijcken:} Software, Investigation, Writing - Review \& Editing. \textbf{Albert Ali Salah:} Methodology, Supervision, Writing - Review \& Editing. \textbf{Floortje Scheepers:} Investigation, Resources, Funding acquisition. \textbf{Marco Spruit:} Conceptualization, Supervision, Project administration, Funding acquisition, Writing - Review \& Editing.

\section*{Acknowledgements}

This work was supported by Utrecht University, the PsyData team from UMC Utrecht, and KPMG n.v. Useful discussions were held with Sebastian Mildiner and Laurens Breij.

\bibliography{references}

\section*{Supplementary Data}
\label{supplementary_data}

\begin{figure}[H]
\centering
\subfloat[ROC-curve on test set institution A]{\label{a}\includegraphics[width=.6\linewidth]{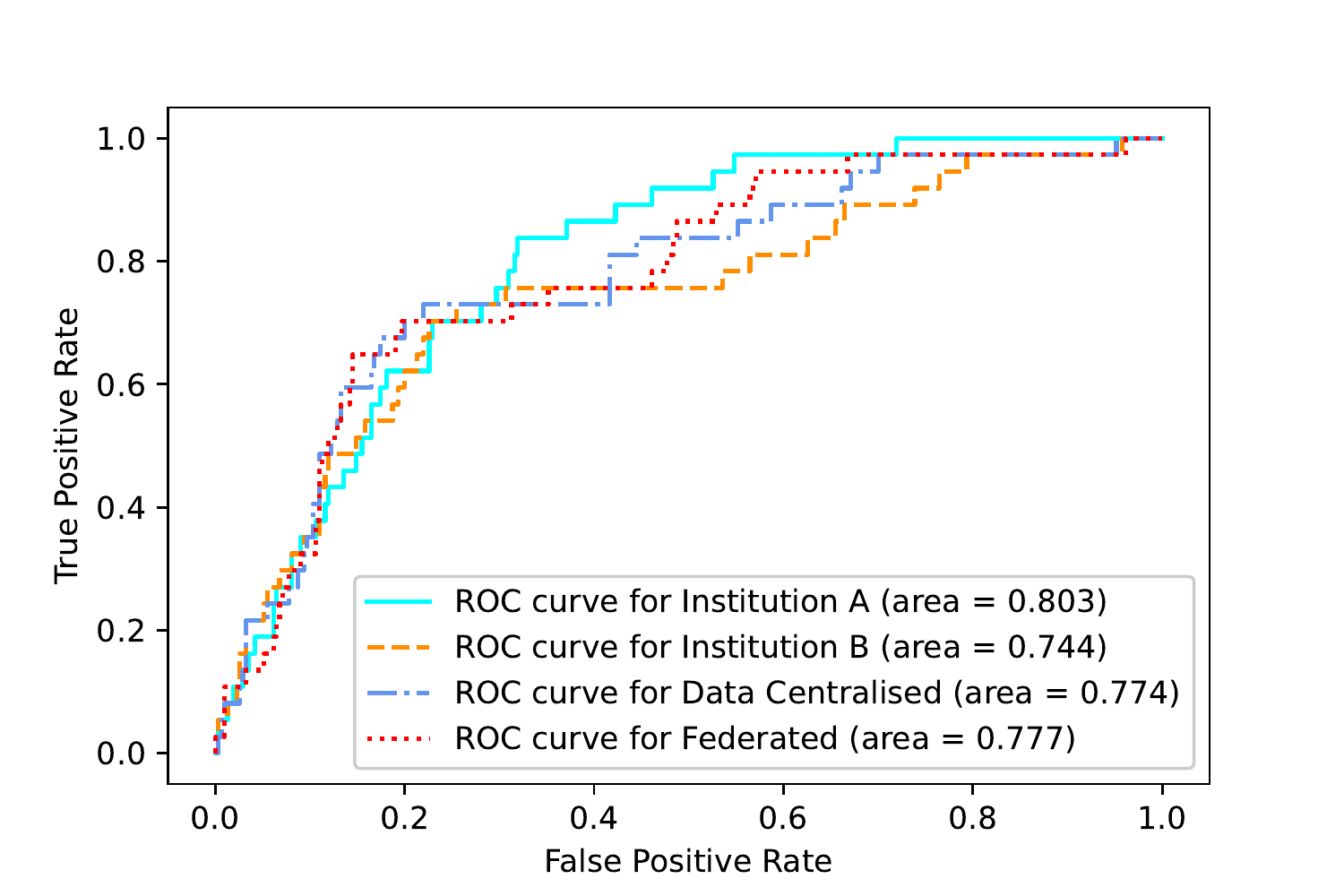}}\hfill
\subfloat[ROC-curve on test set institution B]{\label{b}\includegraphics[width=.6\linewidth]{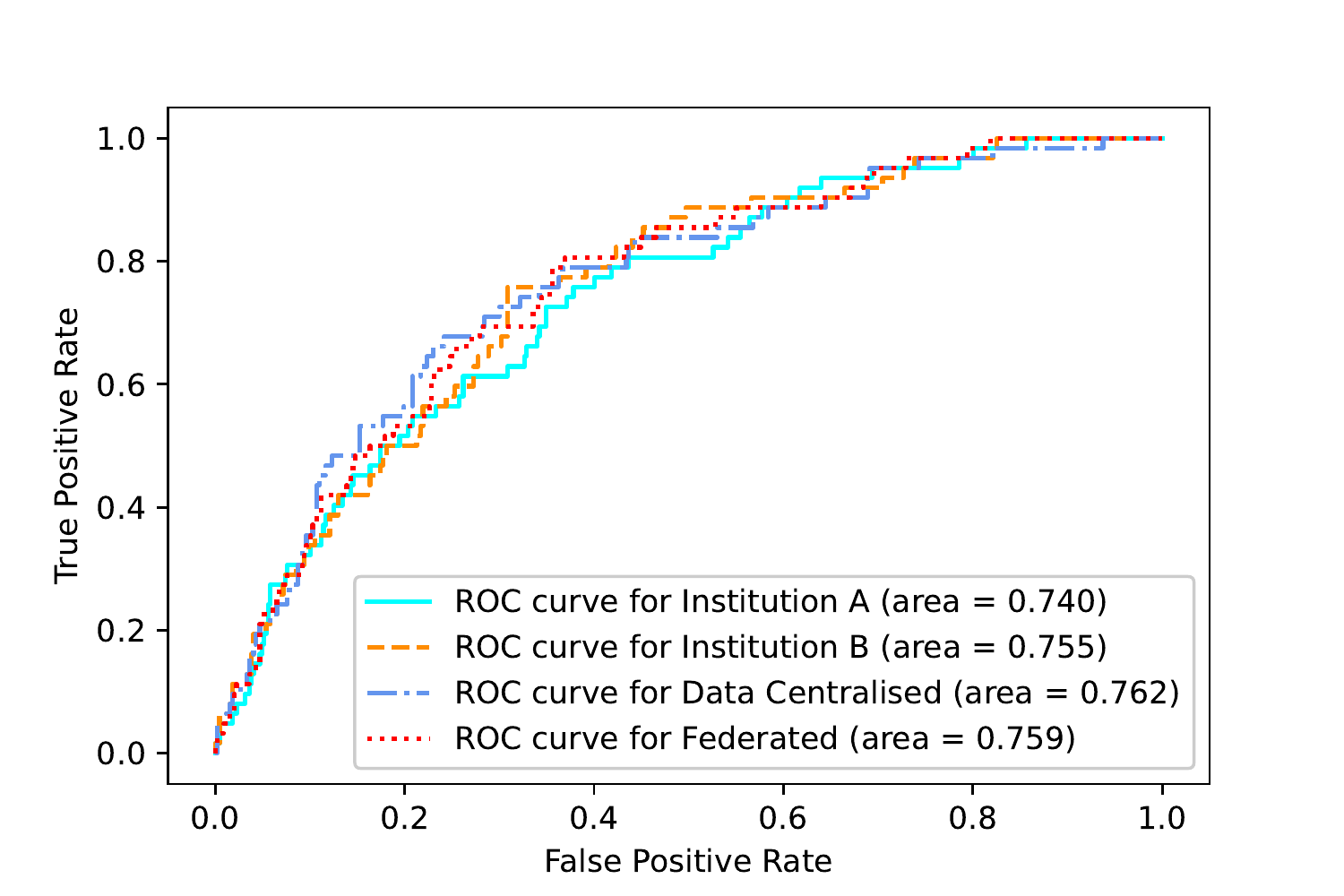}}\par 
\subfloat[ROC-curve on test set combined]{\label{c}\includegraphics[width=.6\linewidth]{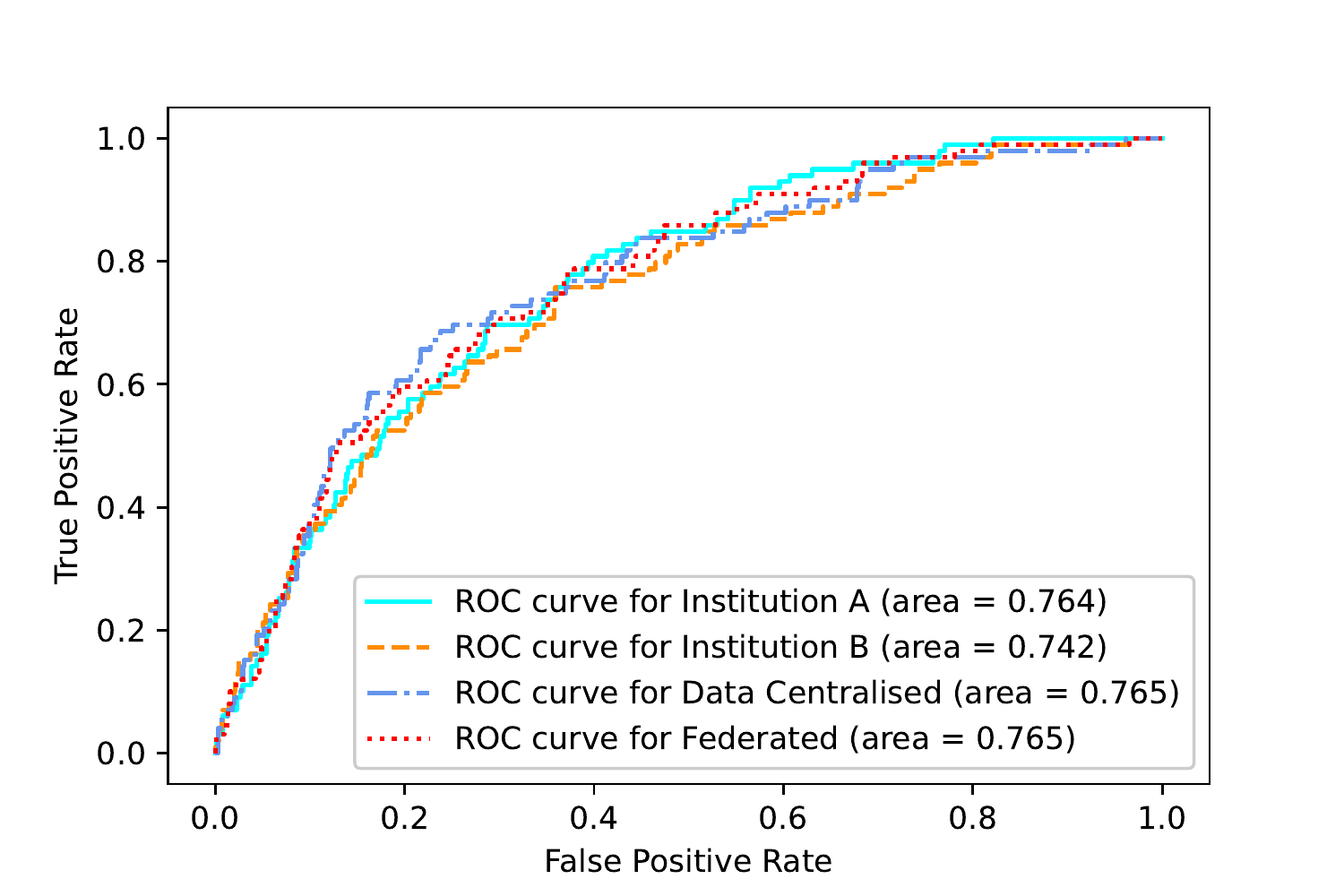}}
\caption{ROC-curves for each treatment's final model tested on each test set. The legend provides the ROC-AUC value.}
\label{fig:roc-curves}
\end{figure}

For each treatment's final model, the receiver operator characteristic (ROC) curve has been computed. Figure~\ref{fig:roc-curves} provides these curves for each test set. 

\end{document}